\documentclass[lettersize,journal]{IEEEtran}
\usepackage{amsmath,amsfonts}
\usepackage{algorithmic}
\usepackage{algorithm}
\usepackage{array}
\usepackage[caption=false,font=normalsize,labelfont=sf,textfont=sf]{subfig}
\usepackage{textcomp}
\usepackage{stfloats}
\usepackage{url}
\usepackage{verbatim}
\usepackage{graphicx}
\usepackage{cite}
\usepackage{xcolor}

\begin{document}

\title{Rise of the Robochemist}

\author{Jihong Zhu$^{1}$, Kefeng Huang$^{1}$, Jonathon Pipe$^{1}$, Chris Horbaczewsky$^{2}$, Andy Tyrrell$^{1}$, and Ian J. S. Fairlamb$^{2}$
\thanks{$^{1}$ The authors are with the School of Physics, Engineering and Technology, University of York}
\thanks{$^{2}$ C. Horbaczewsky and I. Fairlamb are with Department of Chemistry, University of York}\textbf{}
}



\maketitle

\begin{abstract}
Chemistry, a long-standing discipline, has historically relied on manual and often time-consuming processes. While some automation exists, the field is now on the cusp of a significant evolution driven by the integration of robotics and artificial intelligence (AI), giving rise to the concept of the robochemist: a new paradigm where autonomous systems assist in designing, executing, and analyzing experiments. Robochemists integrate mobile manipulators, advanced perception, teleoperation, and data-driven protocols to execute experiments with greater adaptability, reproducibility, and safety. Rather than a fully automated replacement for human chemists, we envisioned the robochemist as a complementary partner that works collaboratively to enhance discovery, enabling a more efficient exploration of chemical space and accelerating innovation in pharmaceuticals, materials science, and sustainable manufacturing. This article traces the technologies, applications, and challenges that define this transformation, highlighting both the opportunities and the responsibilities that accompany the emergence of the robochemist. Ultimately, the future of chemistry is argued to lie in a symbiotic partnership where human intuition and expertise is amplified by robotic precision and AI-driven insight.
\end{abstract}


\section{Introduction}
The field of chemistry, a cornerstone of modern science and industry, has long been characterized by a blend of theoretical insight and practical, hands-on experimentation. For centuries, progress has been driven by the meticulous work of human chemists, whose expertise and intuition have been paramount in navigating the complexities of molecular synthesis, analysis, and characterization. Yet even the most skilled scientist can’t work around the clock: experiments are paused overnight, intermediates are left waiting, and valuable time is lost. Beyond time constraints, manual methods face challenges of safety, speed, scalability, and reproducibility. These limits, once accepted as part of the discipline, are now being challenged by new approaches that promise a step-change in how science gets done.


Automation in the laboratory is nothing new. From the first titration devices \cite{lingane1948automatic} to today’s liquid-handling robots and high-throughput platforms, machines have long been used to accelerate repetitive tasks and improve precision. Alongside gains in productivity, automated systems have also been valued for their ability to log experiments with a consistency difficult to achieve manually, offering a foundation for reproducible data and, more recently, for open digital repositories \cite{kearnes2021open}. Safety has been another important driver: enclosed workstations and interlock mechanisms reduce operator exposure to hazardous materials \cite{omair2023recent}. Nevertheless, despite these advances, most systems remain highly specialised. They are optimised for narrow protocols, dependent on human supervision, and limited in their adaptability to the diverse and dynamic workflows of research laboratories.
\begin{figure}[t]
    \centering
    \includegraphics[width=0.8\linewidth]{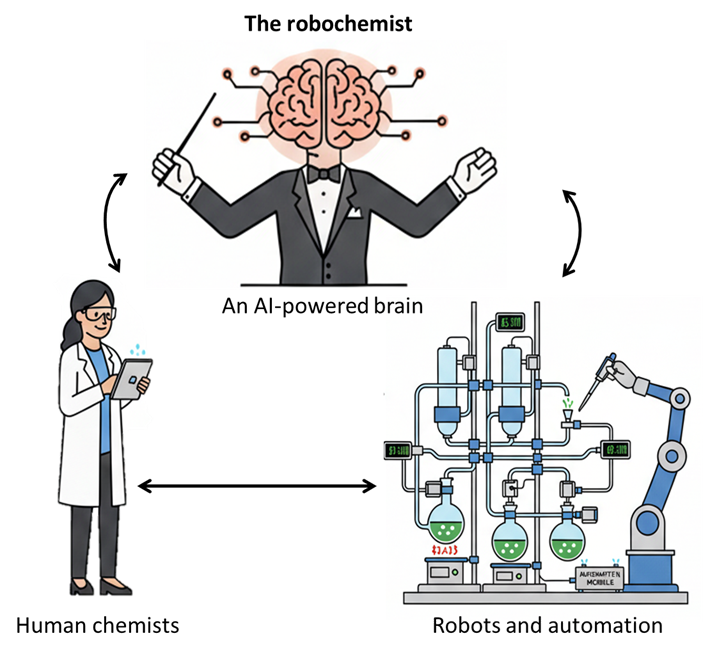}
    \caption{The robochemist vision: an AI-powered brain coordinating the collaboration between human chemists and robots.}
    \label{fig:placeholder}
\end{figure}


Today, robotics is already transforming many aspects of modern life. In households, robots are learning to cook \cite{Sochacki2024Towards}, fold clothes \cite{Garcia-Camacho2020Benchmarking}, and even assist with dressing \cite{zhu2024you}. In industry, they assemble products on factory floors with speed and precision \cite{lai2025roboballet}. These advances highlight how far robotic capabilities have come — and they are beginning to enter the laboratory. In chemistry, mobile manipulators \cite{burger2020mobile} now demonstrate how robots can move through standard laboratory spaces carrying samples between instruments. Some projects \cite{darvish2025organa, yoshikawa2022chemistry} go further, showing how a single robotic arm can perform basic experimental procedures. Just like these pioneering systems, the integration of advanced robotics, sophisticated sensors, and the predictive power of artificial intelligence and machine learning is paving the way for a new era in chemical and molecular discovery — an era of the \textit{robochemist} (see Figure \ref{fig:placeholder}), representing a significant leap beyond simple automation.

This article traces that evolution in three key stages: considering the traditional laboratory automation perspective, an examination of the new technologies that define the robotic and AI era, and a forward-looking discussion of the challenges and opportunities this paradigm shift presents for the future of chemistry and related applied areas.

\section{Laboratory Automation}
The push for automation in chemistry is not a recent phenomenon \cite{olsen2012first}. The mid-20th century saw the emergence of the first generation of laboratory automation systems; devices such as automatic titrators \cite{lingane1948automatic} and clinical analyzers \cite{doi:10.1021/cen-v041n049.p112} became commercially available in the 1950s, enabling mechanized execution of complex, repetitive operations.
However, these systems lacked scientific productivity unless paired with structured experimental designs. Design of Experiments (DoE) methodologies—though originally developed earlier in statistics—became essential tools once such automation systems were in place, providing a principled framework for efficient exploration of multidimensional parameter spaces \cite{bowden2019design, murray2016application}.
Building on this foundation, the 1990s pharmaceutical industry embraced High‑Throughput Experimentation (HTE) by integrating robotics, miniaturization, and parallelization into automated platforms guided by DoE‑inspired strategies—paving the way for rapid chemical discovery and optimization \cite{maier2019early, dar2004high}.
Together, these developments defined the first wave of laboratory automation: mechanical precision powered by systematic experimentation and elevated by high-throughput technologies.

However, the reality of early laboratory automation was more constrained than this promise suggests. These systems were highly specialized, typically designed to execute specific, repetitive tasks at scale. Some examples of these systems includes liquid handling units, high-throughput screening (HTS) platforms, and automated synthesis or purification modules. They process transformative impact on throughput and reproducibility. However, they are usually rigid machines that cannot adapt to changes. In addition, these systems demand constant supervision and maintenance by trained specialists.
In short, they relieved chemists of repetitive manual work but remained limited to narrow functions.

Recognizing these shortcomings, bespoke systems have been developed to target specific domains. For example, the Schlenkputer \cite{bell2024autonomous} automates Schlenk-line chemistry using customized glassware such as filter flasks and NMR adapters, enabling autonomous execution of air-sensitive reactions. Similarly, RoboChem \cite{slattery2024automated} incorporates a purpose-built continuous-flow photo-reactor with integrated in-line NMR and IoT sensors, allowing closed-loop optimization and scale-up of photo-catalytic reactions. These specialized designs demonstrate how tailored hardware can overcome domain-specific challenges, but they inevitably remain tied to their specialized context and still require human oversight.

In parallel, some attempts at generalization have focused on orchestrating workflows across heterogeneous laboratory systems \cite{hamediradtowards}. While these approaches mark an important step forward, they remain constrained by the limitations of commercial automation tools. Most existing platforms are optimized for narrow domains—such as liquid handling or plate-based assays—and do not readily extend to more sophisticated analytical techniques like gas chromatography–mass spectrometry (GC–MS) or liquid chromatography–mass spectrometry (LC–MS). Incorporating such instrumentation often requires bespoke engineering that is impractical for most laboratories \cite{hamedirad2019towards}. In response to these constraints, digital frameworks such as the XDL \cite{mehr2020universal} have been proposed. By providing a machine-readable language for encoding synthetic steps independently of specific instruments, XDL aims to make chemistry protocols portable across platforms. 
By abstracting chemistry into digital protocols, XDL not only makes experiments portable across platforms but also provides a natural mechanism for recording experimental details in a structured way, paving the path toward systematic data capture and reuse.

In this context, automated platforms—particularly high-throughput screening devices and flow chemistry systems—have become powerful tools for generating structured datasets. By logging each step of an experiment with greater precision and consistency than manual record keeping, they produce outputs that can be contributed to open repositories such as the Open Reaction Database \cite{kearnes2021open}. These initiatives aim to capture, standardize, and disseminate experimental knowledge in machine-readable formats, enabling benchmarking, meta-analysis, and AI-driven hypothesis generation across the community. However, a drawback remains: despite the increased use of automation, many workflows still require human intervention, and these steps are difficult to encode digitally. As a result, the records often resemble enhanced laboratory notebooks rather than complete digital traces, with crucial but informal actions—such as small adjustments, setup decisions, or contextual observations—going unrecorded. These omissions can prove critical for reproducibility, highlighting the gap between automated logging and fully transparent experimental records.

Beyond productivity and reproducibility, conventional laboratory automation systems have also been designed with safety in mind. Features such as enclosed liquid handlers, interlock mechanisms, and controlled-atmosphere enclosures help minimize accidental exposure to hazardous substances, thereby lowering the biological risks faced by laboratory operators \cite{lippi2019advantages}. These safeguards underscore that automation is valuable not only for efficiency but also for protecting personnel \cite{omair2023recent}. At a broader level, robotics and automation are increasingly recognized for their potential to make scientific experimentation safer as laboratories evolve toward higher levels of autonomy \cite{angelopoulos2024transforming}. Nevertheless, because such systems still depend on human setup, supervision, and frequent intervention, they cannot fully eliminate the risks associated with handling dangerous reagents or operating complex instruments.

As a result, conventional automation remains highly specialized: effective for large-scale, standardized production, yet far less practical for the diverse and dynamic workflows of typical research laboratories. At the same time, the growing emphasis on structured data highlights both the promise and the challenge of automation: while automated systems are uniquely positioned to capture reproducible experimental records, the lack of universal standards limits their broader impact. Consequently, chemists continue to perform many repetitive and hazardous tasks, even as automation steadily reshapes the way experimental knowledge is produced, shared, and reused.

\section{New Era for Robotics and AI}

The limitations of traditional lab automation—its rigidity, lack of adaptability, and reliance on static protocols—paved the way for a new paradigm in chemical research. 
At the center of this shift is the idea of the \textit{robochemist}: not a single machine, but an ecosystem of interconnected technologies in which robotics and artificial intelligence converge to create flexible and intelligent systems. Unlike their predecessors, these platforms are designed not only to execute tasks, but also to learn from data, adapt to changing conditions, and make real-time decisions.



A central trend in the pursuit of laboratory autonomy is the development of mobile manipulator systems that enable the vision of the self-driving laboratory. In this architecture, experiments are distributed across specialized stations — such as prepare workstation or reaction chamber \cite{li2018air} — while a mobile robot navigates the laboratory to transfer intermediates between stations and coordinate the workflow. This station-based design allows complete experimental processes to be executed end-to-end with minimal human intervention. 
The first landmark demonstration \cite{burger2020mobile} showed how a free-roaming robot, integrated with standard laboratory equipment, could autonomously perform hundreds of photo-catalysis experiments, establishing the feasibility of end-to-end robotic experimentation.
This concept was later extended \cite{dai2024autonomous} to encompass a broader range of chemistry by linking mobile robots with modular synthesis and analysis stations, including LC–MS and NMR, thereby expanding the paradigm beyond single-instrument optimization to more general exploratory workflows. 
These works have dramatically reduced the need for human intervention and made it practically feasible to run fully autonomous experiments. However, they are still constrained by their reliance on conventional laboratory instruments, and often require manual preparation or transitional steps between modules. 
A subsequent study \cite{lunt2024modular} illustrated how this limitation could be addressed by incorporating robotic manipulators directly as stations within the workflow — such as dual-arm or mobile robots performing sample preparation and handling — thus blurring the line between mobile couriers and stationary operators and pointing toward a future where both mobility and dexterity are combined to eliminate remaining human bottlenecks.

Building on the integration of mobile platforms and station-based workflows, a complementary line of research has focused on advancing the manipulation capabilities required for chemistry. Rather than limiting automation to sample transport, these studies investigate how robotic arms can perform core laboratory skills directly. For example, robotic pouring and liquid handling tasks have been demonstrated \cite{yoshikawa2022chemistry}, showing that generic robotic platforms can replicate operations traditionally carried out by human chemists. 
Recent approaches \cite{skreta2023errors,yoshikawa2023large} have further leveraged large language models to decompose experimental protocols into simplified XDL instructions, which can then be executed by path-planning algorithms on robotic manipulators. In this way, what was previously out of reach for traditional automation platforms — the translation of abstract digital representations like XDL into real laboratory actions — can now be approached using generic robotic platforms equipped with algorithmic planning, offering a practical pathway toward realizing the autonomy vision.


Before turning to the broader implications, it is important to note that increasing robotic capability does not, by itself, guarantee complete autonomy. Despite their advances, robots remain limited in areas where human chemists excel — such as applying heuristics, exercising creativity, and adapting flexibly to unexpected conditions. The rise of the robochemist therefore does not signal the replacement of human chemists, but rather the augmentation of their capabilities. Robots contribute precision, endurance, and consistency, while humans provide contextual judgment and problem-solving intuition. Studies \cite{duros2019intuition} have shown that when these complementary strengths are combined, human–robot teams can outperform either working alone. This highlights a future in which laboratory progress depends not on autonomy in isolation, but on the synergy between human ingenuity and robotic efficiency. 
To enable this collaboration, modern systems increasingly rely on intuitive interfaces. Large Language Models (LLMs) and advanced task planners allow chemists to specify high-level goals in natural language, which are then translated into executable robotic workflows \cite{ramos2025review}. For example, a recent study demonstrated how ambiguous instructions can be automatically interpreted and resolved into concrete object-manipulation tasks [3]. This approach frees researchers to focus on higher-level reasoning and decision-making, while robots handle the individual experimental steps.

However, current advances remain far from achieving full autonomy. Robotic controllers are still constrained in scope, as no single strategy today can execute the full spectrum of chemical manipulations. In this transitional stage, teleoperation provides a crucial complement \cite{abramson1970aloha}. By enabling chemists to operate robots from a distance, it not only maintains fine control over delicate procedures but also generates rich demonstrations that go far beyond the predefined parameters recorded by conventional automation. These demonstrations supply the training data needed to develop more generalizable control policies—an approach supported by the success of imitation learning in many other domains \cite{o2024open}. At the same time, the same teleoperation channel can function as a human-in-the-loop interface \cite{Wu2025RoboCopilot:}, allowing operators to intervene when the system encounters errors and enabling the robot to improve through human correction data. Beyond data generation, teleoperation also captures entire workflows in a reproducible format, producing high-fidelity records that can be replayed on identical robotic setups and shared with resources such as open reaction databases. Crucially, the physical separation it enforces between chemists and hazardous materials enhances laboratory safety, approaching the protection offered by fully autonomous systems while still retaining human oversight. In this way, teleoperation bridges the autonomy gap by uniting data generation, reproducibility, adaptability, and safety within a single framework. 

As the robotchemist system progresses, a particularly intriguing frontier lies in coupling material discovery algorithms with robotic execution platforms. Recent advances in large-scale computational prediction have expanded the known landscape of stable inorganic crystals by orders of magnitude \cite{merchant2023scaling}, while autonomous laboratories have already demonstrated the ability to synthesize dozens of previously unreported compounds with minimal human intervention \cite{szymanski2023autonomous}. At the same time, cloud-enabled automation platforms have shown that AI-guided synthesis and characterization can be delivered through remote-access robotic laboratories \cite{pyzer2022accelerating}, and iterative theoretical–experimental paradigms driven by robotic AI chemists are beginning to link high-throughput computation with autonomous experimentation in diverse application domains \cite{zhang2025revolutionizing}. Taken together, these developments point toward a closed-loop framework in which predictions guide synthesis, experimental outcomes refine algorithms, and both evolve iteratively toward increasingly complex objectives. This convergence not only accelerates the pace of materials innovation but also establishes a foundation for genuinely self-driving laboratories capable of tackling problems that were once intractable.

The convergence of AI and robotics marks a shift from rigid automation toward intelligent, adaptable, and collaborative systems. This new generation of robochemists is not merely a tool for executing repetitive protocols, but a genuine partner in the discovery process. By combining advanced learning algorithms, teleoperation for safe and reproducible data generation, and intuitive human–robot interfaces, these platforms extend far beyond conventional automation. They safeguard chemists by keeping them at a distance from hazardous materials, while simultaneously empowering them to pursue higher-level reasoning and creative exploration. Rather than replacing human expertise, robochemists augment it, enabling scientists to tackle more ambitious challenges and accelerating progress toward problems once thought intractable.

\section{Challenges and Opportunities}
The vision of a fully autonomous, intelligent laboratory is compelling, but its realization is not without significant hurdles. While the convergence of AI and robotics has opened up new possibilities, several key challenges must be addressed to unlock the full potential of the robochemist.

One of the most pressing challenges is the absence of standardized, modular hardware and software. On the hardware side, traditional high-throughput automation platforms remain rigid and application-specific, limiting their adaptability to new experiments. Robotic platforms, in contrast, promise greater scalability and flexibility, but they too often rely on bespoke designs or custom-built tooling that are difficult to reproduce across laboratories. To fully realize their potential, robotic systems will need to adopt modular, interoperable components — a kind of chemical “Lego set” — that can be reconfigured to suit diverse experimental needs. Equally limiting is the lack of standardization on the software side. Each group typically develops its own bespoke control software and interfaces, making it impractical for non-specialists to adopt these systems at scale. This fragmentation prevents wider adoption: chemists cannot be expected to master a different interface for every platform. To move toward commercial viability, the community must converge on intuitive, user-friendly, and universally compatible software frameworks.

Another significant technical challenge lies in the perception and manipulation of the robotic systems. While modern robots can handle precise tasks, they still struggle with the variability and unpredictability of a real-world chemistry lab. Small disruptions — such as spilled liquids, irregularly shaped glassware, or misplaced vials — can cause an entire workflow to fail. Beyond these practical issues, there are perceptual challenges unique to chemistry. Many phenomena that are immediately obvious to a human observer — such as a liquid beginning to boil, a gas condensing, or crystals forming in a solution — are subtle and difficult for robots to detect or interpret. Addressing these gaps will require richer sensory feedback, such as haptic and multi-modal sensing, combined with more advanced control and perception policies and AI models capable of reasoning about material transformations and adapting dynamically to unexpected conditions.

The effectiveness of any AI-driven system is only as good as the data it's trained upon. 
In the context of robotic chemistry, the scarcity of suitable datasets remains a major obstacle. While open repositories such as the Open Reaction Database \cite{kearnes2021open} have made progress in sharing structured experimental results, these resources are primarily intended for human interpretation and reproducibility rather than for training AI models at scale. Crucial elements such as multimodal sensor data, contextual annotations of laboratory actions, or comprehensive logs of both successful and failed experiments are often absent. This limits the ability to develop robust models for perception, manipulation, and autonomous decision-making. To unlock the full potential of the robochemist, the opportunity lies in creating richer and more standardized datasets explicitly designed for machine learning and robotic training, ensuring that AI systems can generalize beyond narrow use cases and adapt to the complexity of real laboratories.


Perhaps the most crucial aspect of this new era is the evolving relationship between humans and machines. The challenge is to design systems that are truly collaborative, where the human chemist retains intellectual control and creative freedom while the robochemist handles the labor-intensive and repetitive tasks. This partnership requires user interfaces that are intuitive and allow for seamless communication of high-level goals. Yet a further barrier lies in education: undergraduate training in chemistry still focuses almost exclusively on manual workflows, with little exposure to automation or robotics. As a result, there is often no clear transition pathway between the manual techniques students are taught and the automated systems increasingly used in research and industry. Addressing this gap will be essential if the next generation of chemists is to fully exploit these emerging technologies. The opportunities here are immense, as this synergy could lead to discoveries that would be impossible for either a human or a robot working alone. The future of chemistry lies not in a "lights-out" laboratory, but in a dynamic collaboration that enhances human ingenuity.

\section{Conclusion}
The rise of robochemist is not about an incremental step toward greater laboratory automation but rather a transformative change in the strategy and tactics of chemical discovery. The merging of robotics, artificial intelligence enables laboratories to break out of the mold of rigid special-purpose tools and move toward an environment of flexibility and collaboration. Such systems will not substitute the creativity and judgment of human chemists but rather augment these qualities by adding exactness, repeatability, and stamina.

Going forward, the challenge is not just perception, manipulation, or data-driven control. It is also a challenge of interoperability, safety, and accessibility between laboratories, plus developing the real human scientist of tomorrow in this new paradigm and finding ways to bridge the gap between traditional hands-on training and automated experimentation.
\bibliographystyle{IEEEtran}
\bibliography{ref}

\begin{thebibliography}{10}
\providecommand{\url}[1]{#1}
\csname url@samestyle\endcsname
\providecommand{\newblock}{\relax}
\providecommand{\bibinfo}[2]{#2}
\providecommand{\BIBentrySTDinterwordspacing}{\spaceskip=0pt\relax}
\providecommand{\BIBentryALTinterwordstretchfactor}{4}
\providecommand{\BIBentryALTinterwordspacing}{\spaceskip=\fontdimen2\font plus
\BIBentryALTinterwordstretchfactor\fontdimen3\font minus \fontdimen4\font\relax}
\providecommand{\BIBforeignlanguage}[2]{{%
\expandafter\ifx\csname l@#1\endcsname\relax
\typeout{** WARNING: IEEEtran.bst: No hyphenation pattern has been}%
\typeout{** loaded for the language `#1'. Using the pattern for}%
\typeout{** the default language instead.}%
\else
\language=\csname l@#1\endcsname
\fi
#2}}
\providecommand{\BIBdecl}{\relax}
\BIBdecl

\bibitem{lingane1948automatic}
J.~J. Lingane, ``Automatic potentiometric titrations,'' \emph{Analytical Chemistry}, vol.~20, no.~4, pp. 285--292, 1948.

\bibitem{kearnes2021open}
S.~M. Kearnes, M.~R. Maser, M.~Wleklinski, A.~Kast, A.~G. Doyle, S.~D. Dreher, J.~M. Hawkins, K.~F. Jensen, and C.~W. Coley, ``The open reaction database,'' \emph{Journal of the American Chemical Society}, vol. 143, no.~45, pp. 18\,820--18\,826, 2021.

\bibitem{omair2023recent}
A.~O.~M. Omair, A.~M.~A. Jabbar, and M.~O. Albulushi, ``Recent advancements in laboratory automation technology and their impact on scientific research and laboratory procedures,'' \emph{International journal of health sciences}, vol.~7, no.~S1, pp. 3043--3052, 2023.

\bibitem{Sochacki2024Towards}
G.~Sochacki, X.~Zhang, A.~Abdulali, and F.~Iida, ``Towards practical robotic chef: Review of relevant work and future challenges,'' \emph{Journal of Field Robotics}, vol.~41, pp. 1596 -- 1616, 2024.

\bibitem{Garcia-Camacho2020Benchmarking}
I.~Garcia-Camacho, G.~Alenyà, D.~Kragic, M.~Lippi, M.~C. Welle, H.~Yin, R.~Antonova, A.~Varava, J.~Borràs, C.~Torras, and A.~Marino, ``Benchmarking bimanual cloth manipulation,'' \emph{IEEE Robotics and Automation Letters}, vol.~5, pp. 1111--1118, 2020.

\bibitem{zhu2024you}
J.~Zhu, M.~Gienger, G.~Franzese, and J.~Kober, ``Do you need a hand?--a bimanual robotic dressing assistance scheme,'' \emph{IEEE Transactions on Robotics}, vol.~40, pp. 1906--1919, 2024.

\bibitem{lai2025roboballet}
M.~Lai, K.~Go, Z.~Li, T.~Kr{\"o}ger, S.~Schaal, K.~Allen, and J.~Scholz, ``Roboballet: Planning for multirobot reaching with graph neural networks and reinforcement learning,'' \emph{Science Robotics}, vol.~10, no. 106, p. eads1204, 2025.

\bibitem{burger2020mobile}
B.~Burger, P.~M. Maffettone, V.~V. Gusev, C.~M. Aitchison, Y.~Bai, X.~Wang, X.~Li, B.~M. Alston, B.~Li, R.~Clowes \emph{et~al.}, ``A mobile robotic chemist,'' \emph{Nature}, vol. 583, no. 7815, pp. 237--241, 2020.

\bibitem{darvish2025organa}
K.~Darvish, M.~Skreta, Y.~Zhao, N.~Yoshikawa, S.~Som, M.~Bogdanovic, Y.~Cao, H.~Hao, H.~Xu, A.~Aspuru-Guzik \emph{et~al.}, ``Organa: a robotic assistant for automated chemistry experimentation and characterization,'' \emph{Matter}, vol.~8, no.~2, 2025.

\bibitem{yoshikawa2022chemistry}
N.~Yoshikawa, A.~Z. Li, K.~Darvish, Y.~Zhao, H.~Xu, A.~Kuramshin, A.~Aspuru-Guzik, A.~Garg, and F.~Shkurti, ``Chemistry lab automation via constrained task and motion planning,'' \emph{arXiv preprint arXiv:2212.09672}, 2022.

\bibitem{olsen2012first}
K.~Olsen, ``The first 110 years of laboratory automation: technologies, applications, and the creative scientist,'' \emph{Journal of Laboratory Automation}, vol.~17, no.~6, pp. 469--480, 2012.

\bibitem{doi:10.1021/cen-v041n049.p112}
\BIBentryALTinterwordspacing
``Instruments for clinical,'' \emph{Chemical \& Engineering News Archive}, vol.~41, no.~49, pp. 112--128, 1963. [Online]. Available: \url{https://doi.org/10.1021/cen-v041n049.p112}
\BIBentrySTDinterwordspacing

\bibitem{bowden2019design}
G.~D. Bowden, B.~J. Pichler, and A.~Maurer, ``A design of experiments (doe) approach accelerates the optimization of copper-mediated 18f-fluorination reactions of arylstannanes,'' \emph{Scientific reports}, vol.~9, no.~1, p. 11370, 2019.

\bibitem{murray2016application}
P.~M. Murray, F.~Bellany, L.~Benhamou, D.-K. Bu{\v{c}}ar, A.~B. Tabor, and T.~D. Sheppard, ``The application of design of experiments (doe) reaction optimisation and solvent selection in the development of new synthetic chemistry,'' \emph{Organic \& biomolecular chemistry}, vol.~14, no.~8, pp. 2373--2384, 2016.

\bibitem{maier2019early}
W.~F. Maier, ``Early years of high-throughput experimentation and combinatorial approaches in catalysis and materials science,'' \emph{ACS combinatorial science}, vol.~21, no.~6, pp. 437--444, 2019.

\bibitem{dar2004high}
Y.~L. Dar, ``High-throughput experimentation: A powerful enabling technology for the chemicals and materials industry,'' \emph{Macromolecular rapid communications}, vol.~25, no.~1, pp. 34--47, 2004.

\bibitem{bell2024autonomous}
N.~L. Bell, F.~Boser, A.~Bubliauskas, D.~R. Willcox, V.~S. Luna, and L.~Cronin, ``Autonomous execution of highly reactive chemical transformations in the schlenkputer,'' \emph{Nature Chemical Engineering}, vol.~1, no.~2, pp. 180--189, 2024.

\bibitem{slattery2024automated}
A.~Slattery, Z.~Wen, P.~Tenblad, J.~Sanjos{\'e}-Orduna, D.~Pintossi, T.~den Hartog, and T.~No{\"e}l, ``Automated self-optimization, intensification, and scale-up of photocatalysis in flow,'' \emph{Science}, vol. 383, no. 6681, p. eadj1817, 2024.

\bibitem{hamediradtowards}
M.~HamediRad, R.~Chao, S.~Weisberg, J.~Lian, S.~Sinha, and H.~Zhao, ``Towards a fully automated algorithm driven platform for biosystems design. nat commun 2019; 10: 5150.''

\bibitem{hamedirad2019towards}
------, ``Towards a fully automated algorithm driven platform for biosystems design,'' \emph{Nature communications}, vol.~10, no.~1, p. 5150, 2019.

\bibitem{mehr2020universal}
S.~H.~M. Mehr, M.~Craven, A.~I. Leonov, G.~Keenan, and L.~Cronin, ``A universal system for digitization and automatic execution of the chemical synthesis literature,'' \emph{Science}, vol. 370, no. 6512, pp. 101--108, 2020.

\bibitem{lippi2019advantages}
G.~Lippi and G.~Da~Rin, ``Advantages and limitations of total laboratory automation: a personal overview,'' \emph{Clinical Chemistry and Laboratory Medicine (CCLM)}, vol.~57, no.~6, pp. 802--811, 2019.

\bibitem{angelopoulos2024transforming}
A.~Angelopoulos, J.~F. Cahoon, and R.~Alterovitz, ``Transforming science labs into automated factories of discovery,'' \emph{Science Robotics}, vol.~9, no.~95, p. eadm6991, 2024.

\bibitem{li2018air}
J.~Li, Y.~Lu, Y.~Xu, C.~Liu, Y.~Tu, S.~Ye, H.~Liu, Y.~Xie, H.~Qian, and X.~Zhu, ``Air-chem: authentic intelligent robotics for chemistry,'' \emph{The Journal of Physical Chemistry A}, vol. 122, no.~46, pp. 9142--9148, 2018.

\bibitem{dai2024autonomous}
T.~Dai, S.~Vijayakrishnan, F.~T. Szczypi{\'n}ski, J.-F. Ayme, E.~Simaei, T.~Fellowes, R.~Clowes, L.~Kotopanov, C.~E. Shields, Z.~Zhou \emph{et~al.}, ``Autonomous mobile robots for exploratory synthetic chemistry,'' \emph{Nature}, vol. 635, no. 8040, pp. 890--897, 2024.

\bibitem{lunt2024modular}
A.~M. Lunt, H.~Fakhruldeen, G.~Pizzuto, L.~Longley, A.~White, N.~Rankin, R.~Clowes, B.~Alston, L.~Gigli, G.~M. Day \emph{et~al.}, ``Modular, multi-robot integration of laboratories: an autonomous workflow for solid-state chemistry,'' \emph{Chemical Science}, vol.~15, no.~7, pp. 2456--2463, 2024.

\bibitem{skreta2023errors}
M.~Skreta, N.~Yoshikawa, S.~Arellano-Rubach, Z.~Ji, L.~B. Kristensen, K.~Darvish, A.~Aspuru-Guzik, F.~Shkurti, and A.~Garg, ``Errors are useful prompts: Instruction guided task programming with verifier-assisted iterative prompting,'' \emph{arXiv preprint arXiv:2303.14100}, 2023.

\bibitem{yoshikawa2023large}
N.~Yoshikawa, M.~Skreta, K.~Darvish, S.~Arellano-Rubach, Z.~Ji, L.~Bj{\o}rn~Kristensen, A.~Z. Li, Y.~Zhao, H.~Xu, A.~Kuramshin \emph{et~al.}, ``Large language models for chemistry robotics,'' \emph{Autonomous Robots}, vol.~47, no.~8, pp. 1057--1086, 2023.

\bibitem{duros2019intuition}
V.~Duros, J.~Grizou, A.~Sharma, S.~H.~M. Mehr, A.~Bubliauskas, P.~Frei, H.~N. Miras, and L.~Cronin, ``Intuition-enabled machine learning beats the competition when joint human-robot teams perform inorganic chemical experiments,'' \emph{Journal of chemical information and modeling}, vol.~59, no.~6, pp. 2664--2671, 2019.

\bibitem{ramos2025review}
M.~C. Ramos, C.~J. Collison, and A.~D. White, ``A review of large language models and autonomous agents in chemistry,'' \emph{Chemical science}, 2025.

\bibitem{abramson1970aloha}
N.~Abramson, ``The aloha system: Another alternative for computer communications,'' in \emph{Proceedings of the November 17-19, 1970, fall joint computer conference}, 1970, pp. 281--285.

\bibitem{o2024open}
A.~O’Neill, A.~Rehman, A.~Maddukuri, A.~Gupta, A.~Padalkar, A.~Lee, A.~Pooley, A.~Gupta, A.~Mandlekar, A.~Jain \emph{et~al.}, ``Open x-embodiment: Robotic learning datasets and rt-x models: Open x-embodiment collaboration 0,'' in \emph{2024 IEEE International Conference on Robotics and Automation (ICRA)}.\hskip 1em plus 0.5em minus 0.4em\relax IEEE, 2024, pp. 6892--6903.

\bibitem{Wu2025RoboCopilot:}
P.~Wu, Y.~Shentu, Q.~Liao, D.~Jin, M.~Guo, K.~Sreenath, X.~Lin, and P.~Abbeel, ``Robocopilot: Human-in-the-loop interactive imitation learning for robot manipulation,'' \emph{ArXiv}, vol. abs/2503.07771, 2025.

\bibitem{merchant2023scaling}
A.~Merchant, S.~Batzner, S.~S. Schoenholz, M.~Aykol, G.~Cheon, and E.~D. Cubuk, ``Scaling deep learning for materials discovery,'' \emph{Nature}, vol. 624, no. 7990, pp. 80--85, 2023.

\bibitem{szymanski2023autonomous}
N.~J. Szymanski, B.~Rendy, Y.~Fei, R.~E. Kumar, T.~He, D.~Milsted, M.~J. McDermott, M.~Gallant, E.~D. Cubuk, A.~Merchant \emph{et~al.}, ``An autonomous laboratory for the accelerated synthesis of novel materials,'' \emph{Nature}, vol. 624, no. 7990, pp. 86--91, 2023.

\bibitem{pyzer2022accelerating}
E.~O. Pyzer-Knapp, J.~W. Pitera, P.~W. Staar, S.~Takeda, T.~Laino, D.~P. Sanders, J.~Sexton, J.~R. Smith, and A.~Curioni, ``Accelerating materials discovery using artificial intelligence, high performance computing and robotics,'' \emph{npj Computational Materials}, vol.~8, no.~1, p.~84, 2022.

\bibitem{zhang2025revolutionizing}
B.~Zhang, Z.~Zhu, H.~Li, J.~Cao, and J.~Jiang, ``Revolutionizing chemistry and material innovation: an iterative theoretical-experimental paradigm leveraged by robotic ai chemists,'' \emph{CCS Chemistry}, vol.~7, no.~2, pp. 345--360, 2025.

\end{thebibliography}
\end{document}